\documentclass{article}

\usepackage[english]{babel}

\usepackage[letterpaper,top=2cm,bottom=2cm,left=3cm,right=3cm,marginparwidth=1.75cm]{geometry}

\usepackage{amsmath}
\usepackage{graphicx}
\usepackage[colorlinks=true, allcolors=blue]{hyperref}
\usepackage{ulem}
\usepackage{marginnote}
\usepackage{ifthen}
\usepackage{booktabs}
\usepackage{wrapfig}

\title{\bf Collectionless Artificial Intelligence}
\author{Marco Gori, Stefano Melacci\\
{\small University of Siena --- Siena, Italy}\\
{\texttt{\small marco.gori@unisi.it, mela@diism.unisi.it}}
}

\begin{document}
\maketitle

%\begin{quote}
%Far better an approximate answer to the right question, which is often vague, than the exact
%answer to the wrong question, which can always be made precise\\
%John Tukey
%\end{quote}

\begin{abstract}
By and large, the professional handling of huge data collections 
is regarded as a fundamental ingredient of  the progress of machine learning 
and of its spectacular results in related disciplines. While the recent debate
on how rogue AI may arise seems to be quite controversial because of different 
influential positions, there is a growing agreement on risks connected 
to the centralization of huge data collections, mostly in the framework of machine learning.
This paper sustains the position that the time has come for thinking 
of new learning protocols where machines conquer cognitive skills 
in a truly human-like context centered on environmental interactions. 
%%%%%
{This comes with specific restrictions on the learning protocol according to the \textit{collectionless principle}, which states that, at each time instant, data acquired from the environment is processed 
 with the purpose of contributing to update the current internal representation of the environment, and that the agent is not given the privilege of recording the temporal stream. Basically, there is neither permission to store the
temporal information coming from the sensors, 
nor to store their internal representations, thus promoting the development of self-organized memorization skills at a more abstract level, instead of relying on bare storage to simulate learning dynamics that are typical of offline learning algorithms.}
%\footnote{\color{red}Of course, storing a small buffer with the most recent samples from %the sensors or their agent-computed representations is in principle allowed to support the %development of collectionless agents. This is clearly different from the long-term storage %of data acquired from the sensors/representations, that goes against the collectionless %principle.}}
%%%
%This implies developing the capability of self-organizing the acquired information, avoid a bare long-term storage of the sensory data. 
%%%
%This gives rise to the \textit{collectionless principle} which states that while data is processed at a certain instant with the purpose of contributing to update the current internal representation of the environment, 
%the agent is not giving the privilege of recording the temporal stream to enable the application of  offline learning algorithms. Basically, there is neither permission to store the temporal information coming from the sensors, nor to store their internal representations.\footnote{Of course, this does not imply} 
%%%%
%This implies developing the capability of self-organizing the acquired information, avoiding a bare long-term storage of the information acquired from the environment.
%%%%%
This purposely extreme position is intended to stimulate 
the development of machines that learn to dynamically organize the information 
by following human-based schemes, which naturally rely on the collectionless principle.
The proposition of this  challenge  suggests developing new foundations 
on computational processes 
of learning and reasoning that might open the doors to a truly orthogonal competitive
track on  AI technologies that avoid data accumulation by design, thus offering a framework 
which is better suited concerning privacy issues, control and customizability. 
Finally, pushing towards massively distributed
computation, the collectionless approach to AI will likely reduce the 
concentration of power in companies
and governments, thus better facing geopolitical issues.

\end{abstract}

\section{Introduction}

\begin{wrapfigure}{r}{0.48\textwidth}
    \centering
    \includegraphics[width=0.46\textwidth]{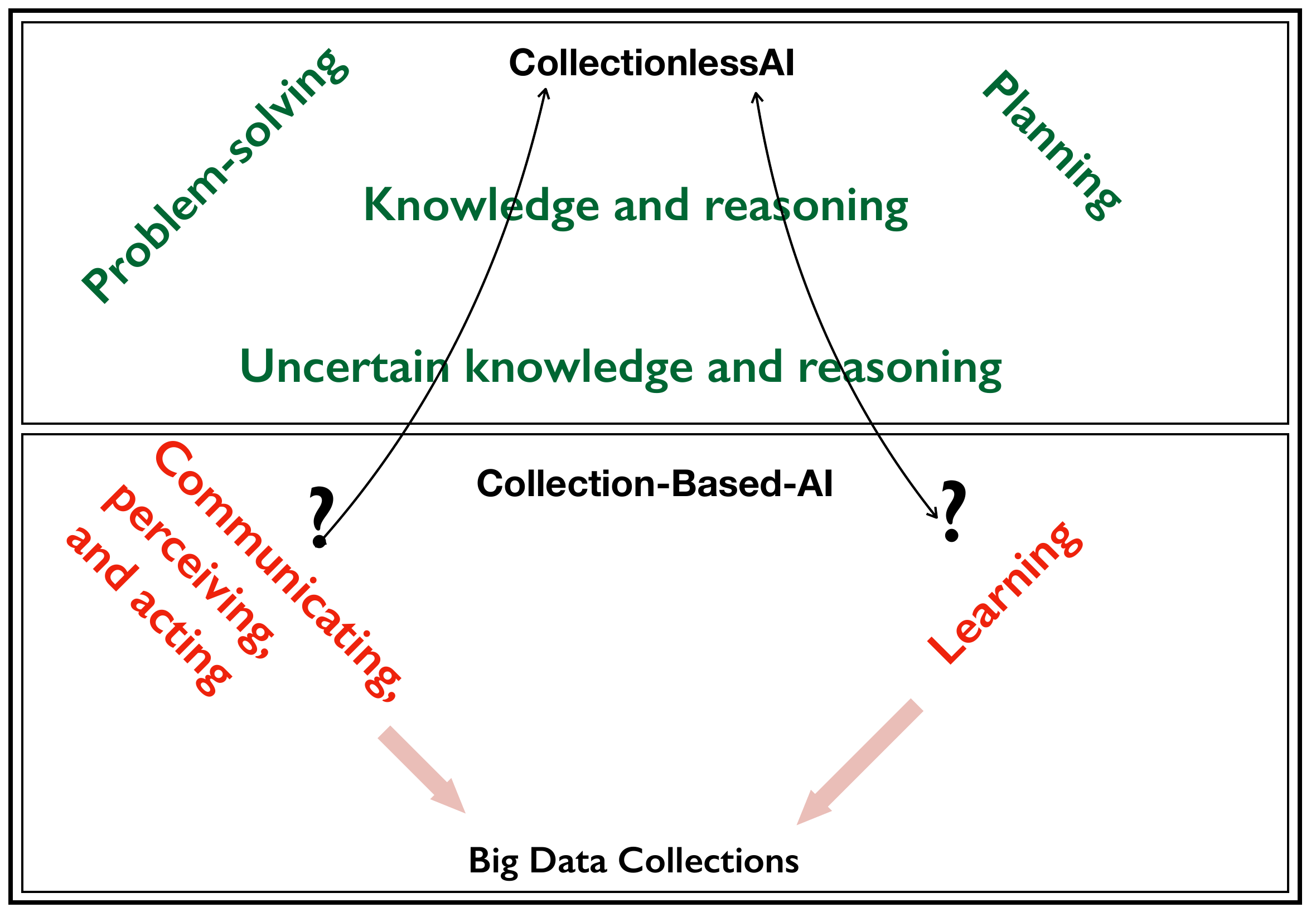}
\end{wrapfigure}
The big picture of Artificial Intelligence that emerges from
by Russel and Norvig~\cite{russel2021} is centered 
around a few classic topics, whose methodologies can, amongst others, be characterized
by the noticeable difference that while ``symbolic AI'' is mostly collectionless,
``sub-symbolic AI'' is currently strongly relying on huge data collections. 
The side figure shows such a difference by providing a rough clustering that is 
induced by commonly adopted methods.  
Interestingly, Machine Learning,
Communicating, Perceiving, and Acting relies mostly on Statistical methodologies
whose effectiveness has been dramatically improved in the last 
decade because of
the access to huge data collections. This has been in fact likely the most important
ingredient of the success of Machine Learning 
that has found a comfortable place under the umbrella of Statistics.
As such, by and large, scientists have gradually become accustomed to 
taking for granted the fact that it is necessary to progressively accumulate increasingly large data collections. 
%It is evident how the quality of data and the easiness of accessing data collections is crucial, %in order to avoid the centralization of the power in the hand of few players as well as to avoid %privacy issues and biases that might easily arise when learning from scarcely controlled (or to %large to control) datasets. 
%Moreover, those solutions that are built on top of the the just mentioned models inherit similar %issues, that are replicated at a possibly smaller scale, for example when specializing the model %to a specific domain.
% Going beyond bare machine learning, mostly having neural networks in mind, data centralization %issues also touches those research directions that bridge learning and reasoning, such as neuro-%symbolic AI, as well as symbolic AI, where the risks/issues related to populating and managing %large knowledge bases are similar to what we discuss here in the case of data collections.
It is noteworthy that even symbolic approaches to AI are based on relevant collections of information, but in this case they are primarily knowledge bases
and there is no data directly collected from the environment. 
When focusing on the difference of information that is stored
a question naturally arises concerning the possibility of exhibiting intelligent 
behavior only thanks to an appropriate internal representation of knowledge.
Clearly, while the knowledge representation typically enjoys the elegance 
and compactness of logic formalism, the storage of patterns apparently leads to 
the inevitable direction of accumulating big data collections. 
However this is indeed very unlikely to happen in nature! 
Animals of all species organize environmental information for their own purposes without collecting the patterns that they acquire every day at every moment of their life.
This leads to believe that there is room for collapsing also the methods that
are used for Machine Learning to the common framework of ``Collectionless AI.''

Throughout this paper, we use the term  ``Collectionless AI'', to identify those 
approaches where intelligent agents do not need to accumulate environmental information. 
As a consequence, when considering Machine Learning, ``Collectionless ML''
identifies those methods which process data at the time in which 
they are acquired from the environment, without storing them. 
We promote the development of dynamic models that adapts over time by 
handling temporal data and learning in a truly online manner. 
The environmental interaction, including the information coming from  humans 
plays a crucial role in the learning process, as well as the agent-by-agent communication. 
We think of agents that can be managed by edge computing devices, 
without necessarily having access to online servers, cloud computing and, 
more generally, to the Internet. 
This requires thinking of new learning protocols where machines 
learn in lifelong manner and are expected to 
conquer cognitive skills in a truly human-like context 
that is characterized by environmental interactions, without storing the information acquired from the environment. 
%??? AI ->ML
%This is significantly different from the majority of popular AI solutions 
%we experienced in the last years. Their outstanding results are not the outcome 
%of following the aforementioned setting, but they come from playing within the largely %established deep learning sandbox, exploiting remarkable (and increasing) computational power. 
%This implies focusing on learning statistically from large, offline-collected datasets, possibly %paired with supervisions, shifting toward the data collection procedure the majority of the %effort needed to setup a new solution, and introducing the issues that we mentioned at the very %beginning of this paper. 
\begin{figure}[!ht]
    \centering
    \includegraphics[width=1.0\textwidth]{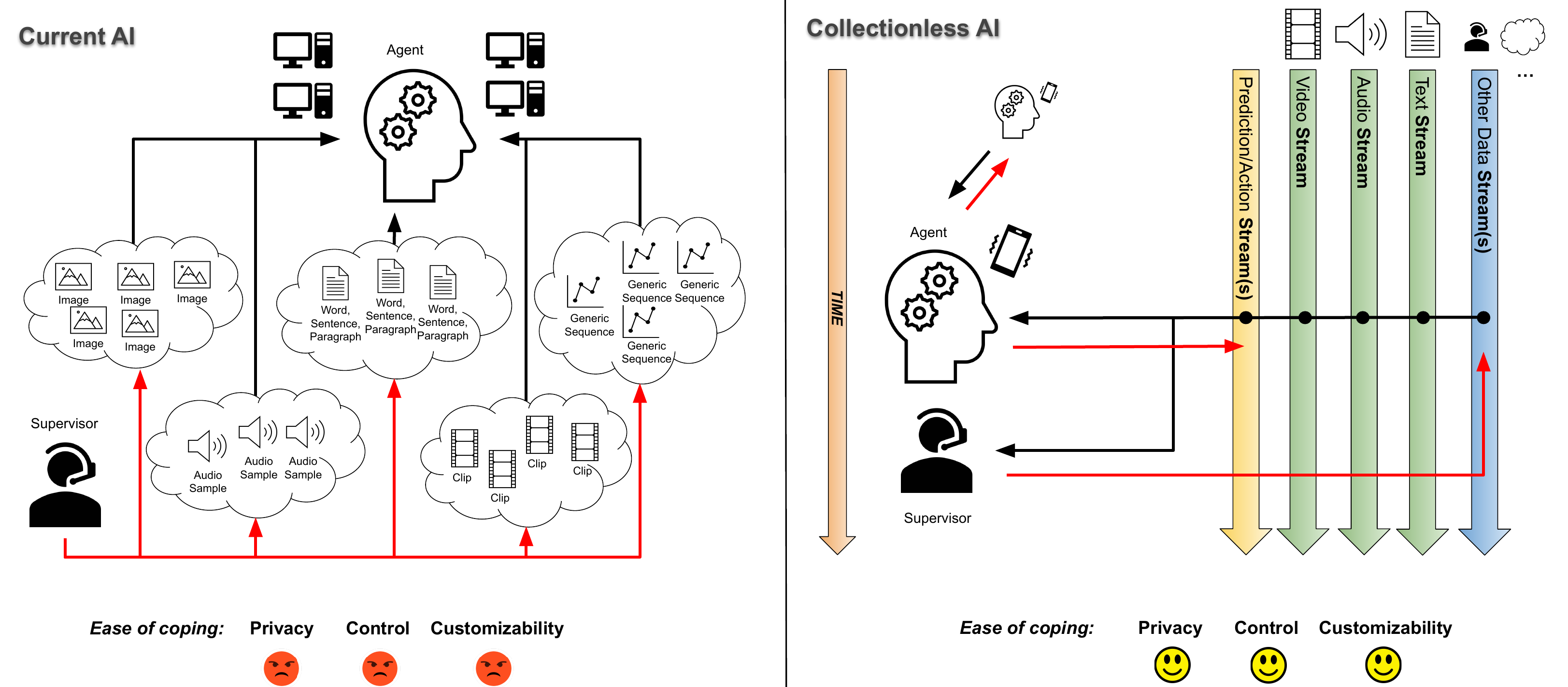}
    \caption{\small Left: Current approaches to Machine Learning focus mostly on learning from large data collections (clouds) that were possibly supervised beforehand. There is no direct/interactive connection between who supervised the data and the agent during the learning process. Learning takes place in (distributed) machines with high computational power. Right: Collectionless AI focuses on learning over time from environmental interactions. The agent ``lives'' in the same environment of the human, with a close interaction and perceives the  information as data streams. 
    Time plays a crucial role, since the way the streamed evolves affects uniformly  learning and inferential processes. The agent can also exploit its own predictions/actions to update its internal state. Computations are distributed over time on edge-computing-like devices, with multiple agents possibly interacting in a social context.}
    \label{fig:generic}
\end{figure}
In Figure~\ref{fig:generic} we compare the protocol of the current Machine Learning 
approaches (left) with the protocol of Collectionless AI (right). 
The former completely depends on data collections (clouds) that were possibly supervised beforehand, and where there is no direct/interactive connection between 
who supervised the data and the learning agent, since all goes through the collections. 
Moreover, learning takes place in (distributed) machines with large computational power 
which supports training procedures which are way shorter that the life of the agent 
after the deployment. Differently, Collectionless AI focuses on environmental 
interactions, where the agent ``lives'' in the same environment of the human, 
with a close interaction. As such, the distinction between learning and test 
that is typically at the core of Statistical Machine Learning 
needs to be integrated with more appropriate methods 
that are expected to rely on ``daily assessments''.  
In this new framework, time plays a crucial role, since the way the 
streamed data evolves affects uniformly learning and inferential processes. 
As a matter of fact, the agent can also exploit its own predictions/actions 
to update its internal state.\footnote{Lifelong/continual learning solutions cover a large number of learning settings. Continual online learning, learning on-the-job \cite{bingliu}, continual reinforcement learning \cite{khetarpal2022towards}, are certainly related to what we discuss in this paper.  %that are the often outcome of sampling in a certain order some data collections that are common to (offline) deep learning benchmarks. 
However, in the supervised continual learning literature, the notion of ``time'' is frequently neglected, and the agent has not an internal state that depends on its previous behaviour/predictions. Continual reinforcement learning follows a specific structure of the learning problem (typical of reinforcement learning, usually augmented with neural networks), while what we discuss here is at a higher abstraction level. In a sense, the ideas of this paper generalize several lifelong-learning-related notions, giving more emphasis to the role of time, to the importance of interactions, and to avoiding the creation of data collections.
%In addition, the paper promotes the ideas of developing dynamic machines that learn how to %organize the information, trying to avoid re-creating the usual conditions for offline-like %learning, by largely storing the sensory information.
} 

Overall, this paper proposes the challenge of introducing
collectionless approaches to learning intelligent skills state in the 
following question: 
\begin{quote}
\it
Learning takes place without accumulation of collections in nature.\\ Is the protocol of Collectionless Machine Learning feasible for gaining intelligent skills?
\end{quote}
From one side, facing this challenge faces the truly scientific issue of 
better understanding the nature of computational processes taking place in 
biology. From the other side, facing this challenge leads to develop AI solutions 
that go beyond the risks connected with data centralization and to
face several questions triggered by those researches that deal with machine learning and that aim at challenging new scenarios that, on the long run, might lead to novel promising technologies more centered around edge-computing-like device, 
without possibly accessing other resources over the network. 
Finally, when facing this challenge once nicely intersects the 
driving directions given in~\cite{depressed}. We do subscribe the authors'
point of view on the picture they give on the  anxious state of many AI scientists
who feel that are not coping with the current pace of AI advancements.
We claim  that Collectionless AI might open the doors to 
important advances of the discipline both from the scientific and technological side. 

%(related to the on-the-job learning in the context of lifelong learning 
%\cite{bingliu})
%}{}
\section{Risks connected with data centralization}
\label{Datacentrisk}
The recent explosion of Large Language Models (LLM) has recently opened 
a strong debate on ``scenarios giving rise to potentially rogue AIs, 
from intentional genocidal behavior of some humans to unintentional 
catastrophe arising because of AI misalignment'' ~\cite{BengioBlog2023}. 
Interestingly, those scenarios are the subject of in-depth discussions 
both in purely scientific fields and in public contexts 
that also involve social and political aspects. 
The positions of internationally renowned scientists, 
including Geoffrey Hinton, Yoshua Bengio, and Yann Le Cun, 
the godfathers of machine learning,  sometimes seem to diverge 
quite significantly.
However, it is important to note how the source of these debates is deeply connected with the progressive exploitation of increasingly large data collections, which require huge financial resources, thus leading to the centralization of information. As will be argued later, this aspect produces undeniable privacy problems as well as producing very controversial geopolitical effects.\\
~\\
\textsc{Data centralization issues}\\
The current massive efforts on the construction of huge data collections 
is strictly connected to the benefits coming from Statistical Machine Learning. 
The progressive accumulation of data has been 
mostly stimulated at the dawn of the web by technologies that have early 
recognized the strategic value of collecting data by massive crawling.  
Search engines companies realized that there were in 
fact great opportunities to access the world wide web information, 
a new black gold, a sort of society's treasure.  
They have become the gatekeepers of our trove of information by
developing extraordinary technologies for accessing the entire Web
and by offering free services of enormous value that were not even remotely 
imagined a few years before.
While blessing those great technological achievements, amongst others, 
some concerns on the Web data centralization 
were expressed in~\cite{WebDragons2007} where search engines are metaphorically 
regarded as \textit{Web dragons}, being them magic, powerful, independent, 
unpredictable, endowed with immense wisdom in oriental folklore, but also 
often portrayed as evil in the west. From one side we can rely on great search services
that make it possible to find successfully stuff on the web. The reverse of the coin is that 
we should feel uneasy about how we have come to rely on search engines so utterly and completely.
The recent explosion of LLM  seems to reinforce the dragon metaphor,
since best technologies are exposed by companies. 
They are likely offering a significant paradigm shift on the type of services 
that are currently being offered. Regardless of the position one might 
maturate on the effectiveness of similar assistants, by and large, it is recognized 
that, because of the impressive impact of associated chatting
services, we are in front of breakthrough in the overall field of AI. 
This holds regardless of the the significant current limitations of LLM 
and associated services that are still far away from getting free of
hallucinations. In any case, just like for the precious Web search services, 
the quality of LLM-based services is strongly associated with the privilege 
of accessing huge data collections, that are exploited by the training algorithms 
of LLM to let the model learn how to embed the processed information into the model parameters and how to drive the dynamics behind the response to the user input.
\\
~\\
\textsc{Privacy and geopolitical issues}\\
The aforementioned considerations are of crucial importance
when considering privacy issues that are easily triggered 
when acquiring data in a private context.
For example, let us consider an agent that processes the data collected from the camera/microphone of a private smartphone.
If the processing does not take place in
the mobile device, there are privacy issues in the way data and interactions are stored.
Unlike the dominant trend of devising computer vision technologies thanks 
to the accumulation of huge visual databases, on the long run, 
learning without visual databases might become an important
requirement for the massive adoption of computer vision technologies. 
This might have a significant
ethical impact and could stimulate novel developments in top-level computer vision labs.
Related discussions hold for voice-based interactions as well as for language processing in 
conversational agents. The current tacitly indisputable collection-based approach to machine learning is not necessarily the only direction for constructing intelligent machines. 
Moreover, while reinforcing the current trend of pushing collection-centered AI 
we implicitly contribute
to create serious geopolitical issues connected with the domain of a few countries which 
can control the technological behavior. On the long run, this has a number of implications in
also in education and in military field.\\
~\\
\textsc{Energy efficiency issues}\\
It is widely known that training LLM, or large Transformers for vision, requires a significant amount of energy, due to the large data collections that are processed by the neural networks. Training can proceed in a distributed manner, for example by splitting the data collections and processing 
%computational burden into different 
different sub-portions of 
data %a given data collection 
in different computational nodes. Basically, most of the energy-related issues are about the attempt to speed-up the training procedure of a single virtual agent operating on a large collected dataset. %Such data might contain redundant, noisy, or not useful information, for the purpose for which the model is trained. 
While these massive pre-training computations  allow us to avoid a complete 
re-training of the machine when focusing on vertical domains,  the overall computational
burden is really heavy.
Moreover, when we restrict attention to domain-specific data collection
and adopt foundational models, still, the training procedures are neither actively-driven by 
the agent nor the supervisor. The collectionless protocol leads to 
interacting with the agent for customizing the teaching processes by 
a more controlled procedure that might avoid providing redundant data/information, 
specializing the training procedure and possibly reducing the computational burden
by focusing only on what is more important for the agent at a certain time instant.
An intriguing perspective is the one offered by multiple collectionless agents exchanging information. Instead of using raw data collections as a mean of exchange, they
can exchange more concise pieces of information that depend on the 
members of the conversation. This information can specifically be crafted 
to help them to speedup their learning process. Distributed computations with brief/targeted communications among agents might reduced the energy consumption needed to develop artificial agents.

However, it is worth mentioning that a reliable comparison on energy consumption 
can hardly be carried out. The main reason is that while collectionless AI is 
saving the massive consumption located in a few big computational centers, 
the diffusion of collectionless machines leads to spreading the consumption 
in a massively distributed way. Their availability would likely give rise to 
intensive distributed computation in thin clients. 
While the comparison on energy consumption is arguable, it has been claimed that
most of current top level machine learning technologies are attaching
learning problems that formulated in such a way that their computational 
complexity is  significantly higher than solving the associated problems in the
temporal dimension~\cite{deeplearningtosee}. 
\\
~\\
~\\
\textsc{Limited control, customizability, and causality}\\
Dedicating several resources in collecting, cleaning, and managing large data collections 
has a clear impact on the expected performance of AI solutions. 
Of course, data might be affected by biases on their content that might be hard to filter out, 
or they might contain inappropriate material. 
As long as the size of the collection increases these issues get more and more serious.
Moreover, controlling data can set expectations on what an artificial agent could 
actually learn, but most modern machine learning solutions do not give any guarantees 
on the behavior of the agents once it has been developed. 
%The major issues behind explainability can hardly be solved by simply maintaining 
%large data collections. 
Whenever the supervisor is only an external entity 
that labels data without being involved in the actual environmental 
interaction, we likely miss important cues.
Differently, the progressive interaction with the environment paves the way of a more controllable AI.

The recent progresses in self-supervised learning mitigated the need of 
manually labeling large training sets, even if labeling still remains 
a significant challenge when specializing the model to vertical domains.
Moreover, in many applications we have been experimenting the benefits 
of transfer learning which relies on pre-trained models 
whose role in fields like computer vision and natural language 
processing has been impressive. On one hand, this speeds-up the 
development and it usually yields improved performances; on the other hand,
pre-trained models inherit the biases induced by the data they were trained on. 
%Even when dealing with transfer learning-based solutions, 
%a key issue with the labeling process is the one of providing labels to
%those examples that are expected to carry the most 
%important pieces of information for the learning task. 
%In fact, measuring the labeling quality simply counting the number of supervised examples
%should not be the only indicator to use when analyzing the collected data. Learning online and %in an interactive manner reduces/avoids the needs of depending on external models or data, and %favours the control on what the model is actually learning, strengthening the connection between %the agent and the supervisor. %, and it does not necessarily require to store the information %exchanged in the agent life. 
%
Finally, the machine learning agents involving perceptual tasks
that are developed by learning on large collections where the 
temporal dimension of the information is missed can only partially 
capture the causal structure of the prediction.

\section{Collectionless AI: features of the learning protocol} 
The evolution of machine learning based on large data collections is mostly due to the 
massive exploitation of classic statistical-based computational models which assume that 
the agent behavior in the real-world environment (test data) is gained after a massive training
over the training set. 
Clearly, the underlying ambition is to acquire data collections as large as 
possible for better estimating the probability distributions. 
A radically different perspective emerges as we think of machines that acquire 
cognitive skills without accessing previously stored data collections, but simply by  
environmental interactions, where the sensory information is immediately processed 
and agent-to-human or agent-to-agent exchanges take place \cite{canmachines}.
In nature, animals do not rely on data collections, but process information as time goes by and create (and update) an appropriate internal representation of the environment. It is in fact the interaction with the environment which allows them to be in touch with the treasure of information, which enables the growth of their cognitive skills.
%Animals  don't  store the information  acquired at a sensory level from sounds, video, odors %and touch. They construct a gradual compact representation of the information which is stored in %their brain in synaptic connections. 
While the connectionist wave has already offered an intriguing interplay 
between the biological and artificial neural neural networks, 
we are still struggling in understanding how  we can create internal representations 
that don't need to access data collections, that is the main focus of Collectionless AI.  
When considering the spectacular results mostly in deep learning, at a first glance, 
we  might regard the proposed Collectionless AI challenge as quite ridiculous. However, 
from one side, because of the way intelligence emerges in nature, 
it is of interest in itself from a truly scientific point of view. 
Moreover, in addition to its importance in the science of AI, we also advocate 
its potential impact for the enabling of a truly different type of AI technologies, 
which could have a dramatic impact in the society.     
The proposed ``restrictions'' on the tools that can be used to develop intelligent agents can 
lead to the activation of a new wave of artificial intelligence pushed by methods somewhat orthogonal to what has been dominating the current state of the art. 
\iffalse
Instead of conceptualizing lifelong learning within the usual offline-learning-designed deep learning sandbox, 
we propose to consider it as an independent problem, that must be handled with appropriate and maybe not-already-available approaches and, to a certain extent, tools. Of course, this does not imply neglecting the already available material, that is definitely precious, but it is intended to avoid doing the mistake of limiting the scientific thought in this research direction due to constraints introduced by what is already available due to the experience gained in other popular problems.
\fi 
\begin{table}
    \centering
    \begin{tabular}{l p{0.8\textwidth}}
    \toprule
    \textbf{} & \textbf{Collectionless AI: Rules of the Game}\\
    \midrule
    1. & \texttt{Data collections are banned} \\ 
    & \footnotesize{Data collections can only be accessed as sequential sources, but
    they cannot be stored for subsequent training processes.
    The sequential access assumes that data collections are presented as collections
    of ``temporal tokens'', so as one can access libraries and any 
    enabled sources of information, but the associated processing must be on-line.}\\
    \\
    2. & \texttt{No permission to store temporal information coming from the 
    sensors for subsequent usage.} \\
    & \footnotesize{Collectionless AI agents are not allowed to store 
    temporal information coming from sensors due to their environmental interaction.
    They can only process the information available at the time of the processing
    and, therefore, they need to produce an appropriate organization of their
    interaction. Data storage from sensors is not allowed for subsequent usage 
    normally driven by statistical processes. 
    {Only ${\rm O}(1)$ space complexity is allowed for buffering 
    sensorial data.}}\\
    \\
    3. & \texttt{No permission to store temporal information developed in internal representations.}\\
    & \footnotesize{Collectionless AI agents cannot store 
    sequences composed of patterns collected from any part of their internal
    representation. Hence, the above restriction to store environmental interactions
    coming from sensors is also extended to the produced internal representations. {Only ${\rm O}(1)$ space complexity is allowed for buffering 
    sensorial data.}
    }\\
    \bottomrule
    \end{tabular}
    \caption{The rules that define the development of learning in the
    framework of Collectionless AI. {Let $[0,n]$ be the temporal 
    horizon defined by $n$ samples. A general learning algorithm can
    stores ${\rm O}(f(n))$ samples. The collectionless principle states
    that only ${\rm O(1)}$ samples are allowed, which promotes 
    solutions based on thin clients.}}
    \label{tab:rules_hard}
\end{table}

\iffalse
    No permission to store the temporal information with the aim of building and possibly reusing data collections to reproduce offline-learning-like procedures. This includes sensory data, latent representations, model outputs, or interactions with other agents/humans. Of course this does not mean that no memory can be used at all to help the learning procedure, but the memory size must be constant with respect the lifespan of the agent
\fi

Table~\ref{tab:rules_hard} describes the Collectionless AI protocol by 
stating the \textit{rules of the game}. Agents operating with this protocol 
must follow three fundamentals rules which dramatically restrict their 
development. The dominant role of large data collections is moved to 
the strength of the permanent environmental interaction. Table~\ref{tab:rules} provides more details on some ingredients involved in collectionless AI.
\begin{table}
    \centering
    \begin{tabular}{l|p{0.7\textwidth}}
    \toprule
    \textbf{Feature} & \textbf{Role in Collectionless AI}\\
    \midrule
    \texttt{Environment} & \small We can think of several scenarios. The agent could share the environment with humans, being exposed to same sensory data that the human would process (audio, video, etc.). Multiple agents could share the environment and interact. 
    Accessing third-party collections (libraries of books, movies in a cinema, datasets from the web, $\ldots$) is possible provided that agents are provided with data streams
    and that they undergo the same computational scheme as for other sensory data.
    Third-party collections are accessed under ordinary permission protocols. \\   
    \\
    \texttt{Time} & \small 
    The role of time in machine learning has taken on many forms and facets over the past few years. However, it is important not to confuse time, which is ordinarily the protagonist of physical processes and, consequently, of the environmental interactions that take place in nature, with various of its relatives already conceived for the treatment of sequential information. For example, ``time'' in Hidden Markov Models or in recurrent neural networks, as well as in reinforcement learning has clear relationships with time as a physical entity, but also evident differences. It is common to consider the computational time, even with its expressions in the field of computational complexity, but this is not strictly related to the time of the environment. In the protocol that drives Collectionless AI, intelligent agents, both natural and artificial, actions are marked by physical time, which remains clearly distinct from the ``internal time'' required for their computation.
    It is in fact the proposed collectionless approach which which gives time the leading role.\\
    %The agent is expected to exploit the natural development of the sensory information to %improve its learning skills. 
    %It must be able to process not only instantaneous snapshots from the sensors, but also to %make-predictions-on/learn-from series of measurements the way they are provided over time. 
    \iffalse We think of agents that do not depend on a sliding window and that are able to dynamically update their internal state in an online manner. There are two degenerate instance that, for simplicity and with the aim of being more general, we also consider. The first one is when inference is instantaneous and does not depend on what happened before. The second one is when the storage includes a small sliding-window buffer. Again, they are not the primary focus of Collectionless AI, but they can be important in the transition toward the agents discussed in this paper. 
    \fi
    \\
    \texttt{Learning} & \small In the case of neural networks there is already abundant evidence of the benefits coming from transfer learning and, recently, from the pre-training in 
    Large Language Models. The collectionless AI protocol significantly emphasizes the use of genetic structures that are formed over generations of agents. In fact, the difference of Collectionless AI agents which grow up in different environments turns out to be the primary characteristic that enables the strength of evolutionary processes. Cooperation is also the other distinctive point that is triggered by the  protocol. Indeed, it is the diversity of the agents combined with their uniformity of ``species'' that stimulates the communicative processes.
    \\
    \bottomrule
    \end{tabular}
    \caption{Collectionless AI: main features of some of the involved ingredients.}
    \label{tab:rules}
\end{table}

Here we promote a truly human-like environmental interaction  which assumes that any sensory information is potentially processed and that the environment is in common between humans and machines. We think of processing of any type of perceptual information coming from any source. Hence, intelligent agents are expected to process visual streams as well as voice coming from  both natural and artificial sources. Moreover, we assume that an agent can, just like humans, access textual information, for example from a library or from the Web, but that, just like for perceptual information, the processed data are not stored. While interactions with humans and the exploration of sensory data inevitably take place at a human-like pace, processing information such as reading a book from a library, watching a pre-recorded movie, can take place at a faster rate, compatibly with the speed at which the data is provided and the processing speed of the agent. However, it is still intended to happen in an online-learning manner, even if it opens to the natural capability of performing multiple passes (multiple readings, watching a move multiple times, etc.). As already anticipated, we also think of multiple agents exchanging information among them, not using raw data collections as a mean of exchange, but more concise pieces of information that depend on the partners of the conversation. Finally, agents are both expected to have a passive role in receiving supervisions and an active behaviour in asking from them.
\iffalse
As we move towards this new direction a number of questions emerge in different fields of 
AI, ranging from video processing \cite{deeplearningtosee} to natural language understanding and related fields.
%
\begin{itemize}
\item Can machines learn to see without offline-processed large visual databases?
\item Can we devise solutions for acquiring visual skills from the simple on-line processing of video
signal and ordinary human-like scene descriptions?
\item Couldn't it be the case that we have been working on a problem that is, from a computational
point of view, remarkably different and likely more difficult with respect to the one offered by
Nature? 
\end{itemize}
\fi
\\
~\\
\textsc{Time as the protagonist of learning}\\
% O(1) space requirements
Sensory information is intrinsically characterized by a natural temporal development of
the data. For example, the case of vision is an idiomatic example that concretely tells that we learn from data that follows a precise temporal order, thus it is not i.i.d., and the learning process naturally proceeds over time. In other words, we do not learn from a huge dataset of shuffled images \cite{deeplearningtosee}. %Supervision does not consist of a detailed labeling of each single piece of information that is acquired, since the way we learn is not only due to a passive exposition to (eventually) pre-buffered data. 
In nature, animals gain visual skills without storing their visual life.
%\begin{itemize}
%\item 
Why can they afford to see without accessing a previously stored visual database? 
%\item 
Is there a specific biological reason that cannot be captured in machines? 
%\end{itemize}
This paper sustains the position that those skills can be likely gained also by machines, once we face the challenge of learning to see without using data collections, exploiting the natural development of the sensory information over time. This argument is supported by the belief that also in nature visual skills obey information-based principles that biology simply exploits to achieve the objectives required by living in a certain environment. We suggest thinking of machines that live in their own visual environment and adopt a natural learning protocol based on the continuous on-line processing of the acquired video signal and on ordinary humans interaction for describing visual scenes.
We can depart from the idea of labeling pre-buffered data to actively interacting with the
artificial agent that is being trained, by exposing it to a form of supervision that depends on the time in which it is provided. In this context, the agent can take the initiative of asking for a supervision and also the (human) supervisor can take the initiative of interacting. This affects dramatically the current labelling procedures, since the interaction takes place in a truly human-like context, thus providing labels that are more appropriate at a specific time instant. 

The agent must be able to process not only instantaneous snapshots from the sensors, but also to make-predictions-on/learn-from series of measurements the way they are provided over time. In the most extreme case, the agent should not depend on a sliding window of hard-to-define length, and be able to dynamically update its internal state in an online manner. Models that make prediction on the current input sample(s), without depending on the outputs/predictions of the previous instants are degenerate instances of what we just described (think about plain image classification). Another somewhat special case is the one in which a small fixed-size buffer is handled  as sliding-window to store the recent sensory data (there might be dependencies among consecutive windows, or not). This is another degenerate instance of what we described so far, and not the primary
focus of Collectionless AI, even if it might have a role in the transition toward
the agents discussed in this paper.
\\
~\\
\textsc{The new scenario}\\
The previous discussion on Collectionless AI suggests a new scenario especially in the evolution of Machine Learning. The features of some of the main ingredients are summarized in Table~\ref{tab:rules}, while the resulting scenario is sketched in the items below.
\begin{itemize}
    \item \textit{Diversity and intelligence: The opposite direction of LLM}\\
    In Section~\ref{Datacentrisk} we have discussed the risks connected with
    data centralization. The recent explosion of LLM is mostly the outcome 
    of the approach of data centralization. This papers suggests following
    exactly the opposite direction. While an orthogonal guideline is 
    highly risk in terms of short-term results, the emergence of 
    intelligent agents that express remarkable differences because of the
    different environments they interact with turns out to be a
    precious ingredient for facing the issues addressed in Section~\ref{Datacentrisk}.
    \iffalse 
    The data collections we discussed so far are large databases of examples that try to summarize all the possible sensory information to which the agent was exposed during its life, or that are large enough to let the agent develop pretty generic and informed representations. When moving to the context of lifelong learning, storing the whole sensory data seen so far, or a subsample that is function of the length of the temporal horizon, yields another instance of data collection. Differently, storing a few information to help the learning process, in a very limited way and that do not depend on the length of the lifespan of the agent, does not fall within our notion of data collection. Of course, we push the idea of machines that learn how to self-organize the information over time.
    \fi
    \item \textit{More natural adaptation to personal interaction: the slowdown challenge}\\
    The Collectionless protocol is better suited for the adaption to personal interaction.
    The development of computational models that adhere to the protocol will likely 
    require a lot of time. On the one hand this can bring fresh air to research, with
    renewed emphasis on academic research~\cite{depressed}.
    On the other hand it can contribute to slowing down the current concentration on studies and developments aimed at reinforcing the paradigm of generative intelligence at the basis of LLMs.
    \item \textit{The trump card} \\
    The restrictions dictated in Table~\ref{tab:rules_hard} appear artificial and, above all, they seem to limit the potential of artificial intelligent agents. This is certainly true, since these ``monastic renunciations'' of the use of memory for collecting data, which represents one of the distinctive aspects of computers, are strictly followed. However, a more careful analysis reveals a much more complex picture. In particular, the renunciation of data collections is balanced by the permanent sequential information processing. The loss of collections can therefore be recovered and potentially overcome by the primary role of time described above. In this scenario, no organizational effort driven by the funding of large companies currently involved in AI can lead to the accumulation of data collections in quantities comparable with what could emerge from the Collectionless AI scenario. This is because the growing number of active thin clients, that continuously acquire information, could soon overturn the privilege to process large collections. 
    \item \textit{The same umbrella for symbolic and sub-symbolic AI}\\ 
    %\item Contribution to moving back to academia 
    As pointed out in the introduction, symbolic approaches to AI mostly 
    reflect the collectionless philosophy. They creation of useful data
    structures (e.g. think of the algorithms at the basis of Problem Solving
    and to knowledge bases) concerns a symbolic level, but it does not involve
    data collections on environmental events. The development of Machine Learning
    theories driven by the collectionless philosophy leads to an interesting
    unification of symbolic and subsymbolic AI. To sum up, machines are allowed to 
    represent knowledge that can be expressed either in symbolic or subsymbolic form
    (e.g. weights of neural networks), whereas the access to data collections
    on environmental events is forbidden. 
\end{itemize}

%\section{Formulation of learning and reasoning in the temporal dimension}
\section{Benchmarks in Collectionless Machine Learning}
Machine Learning has  strongly benefited from the massive diffusion of
benchmarks which, by and large, are regarded as fundamental tools for performance evaluation. 
Alternatively, just like humans, machines can also be expected 
to ``live in their own  environment'' and they can be evaluated on-line \cite{canmachines}.
In such a framework, we need an in-depth re-thinking of nowadays benchmarks.
They bears some resemblance to the influential testing movement in psychology which has its roots in the turn-of-the-century work of Alfred Binet on IQ tests~\cite{Binet1916}.
Both cases consist in attempts to provide a rigorous way of assessing the performance or the aptitude of a (biological or artificial) system, by agreeing on a set of standardized tests which, 
from that moment onward, become the ultimate criterion for validity. 
On the other hand, it is clear that the skills of any agent can be quickly evaluated 
and promptly judged by humans, simply by observing its behavior. 
How much does it take to realize that we are in front of person with visual deficits? 
Do we really need to accumulate tons of supervised examples for assessing the quality of
an artificial agent? 

Scientists in machine learning could start following a sort of en plein air movement \cite{gori2015plein}.\footnote{Here, the term is used  to  mimic 
the French Impressionist painters of the 19th-century and, more 
generally, the act of painting outdoors.} This term suggests that intelligent agents could be evaluated by allowing people to see them in action, virtually opening the doors of research labs. The huge impact of LLM to non-scientific audience is largely due to the availability of a free and simple user interface that every registered user can exploit to submit queries and evaluate the result. This suggests how a massive online active evaluation, opened to a wide audience, can be a viable path for qualitatively evaluating virtual agents that progressively learn. The quality of the same agent can be evaluated at different stages of its evolution, analyzing progresses and regressions. Of course, like any crowd-sourcing-based scheme, this must be carefully ruled to avoid spam. The spirit of this way of evaluating agents is also in line with those machine learning benchmarks that expose existing technologies to ``wild'' settings, different from those in which they have been trained. As a matter of fact, everybody can provide the operating conditions that are considered to be appropriate for a target agent and to accommodate the curiosity of whoever is evaluating.
Now we present a few benchmarks that nicely reflect the 
spirit of collectionless Machine Learning. They are mostly concerned to identify the spirit of the the collectionless Machine Learning protocol and are not supposed to propose a competition framework  with the corresponding  well-established benchmarks of Machine Learning. 
We propose a few examples mostly in the area of generative-AI, though the
collectionless Machine Learning protocol can be formulated also for 
classic classification tasks.

\begin{figure}[htbp]
  \begin{minipage}{0.5\textwidth}
    \centering
    \includegraphics[width=\linewidth]{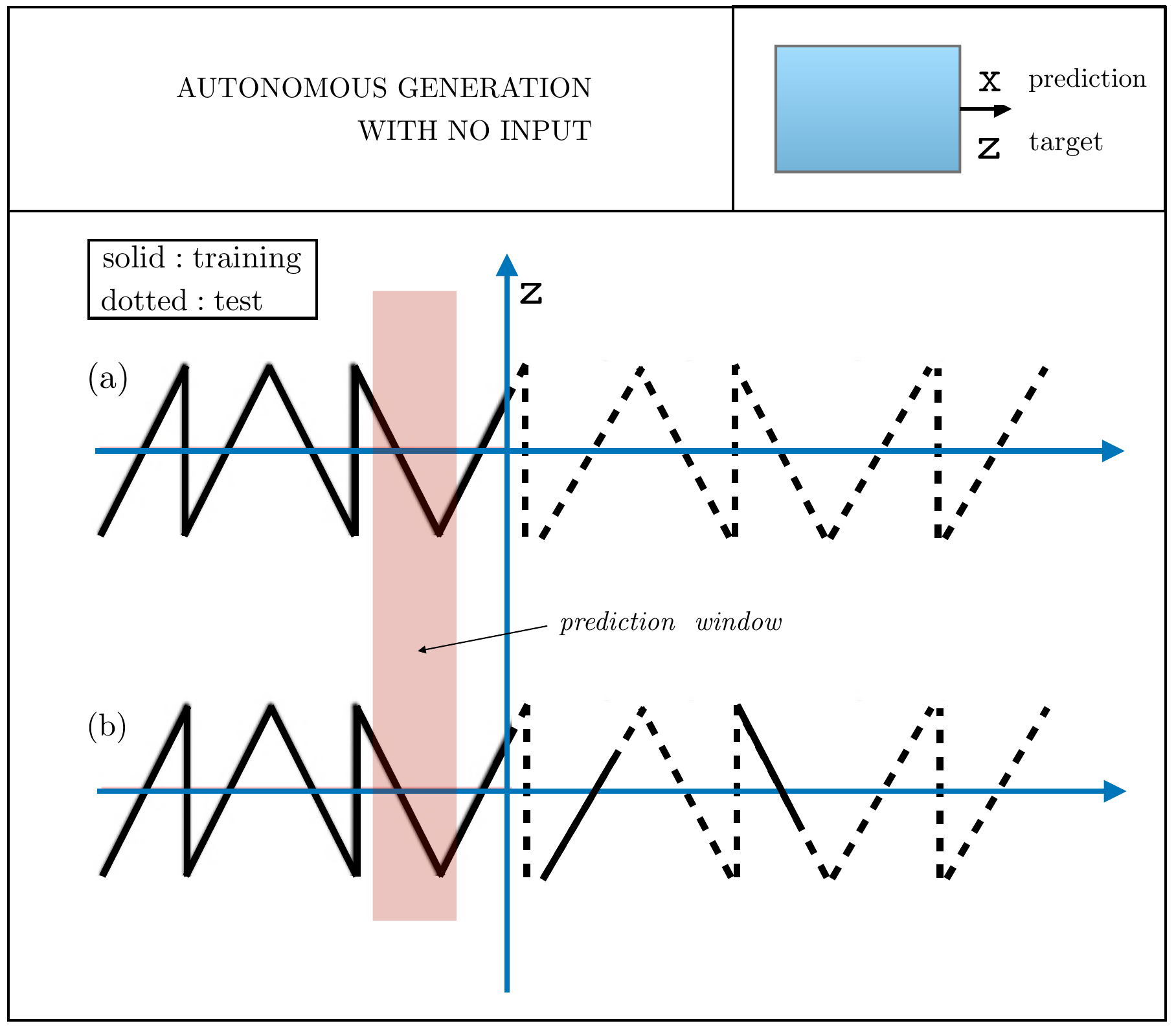}
    \label{fig:figure1}
  \end{minipage}%
  \begin{minipage}{0.465\textwidth}
    \centering
    \includegraphics[width=\linewidth]{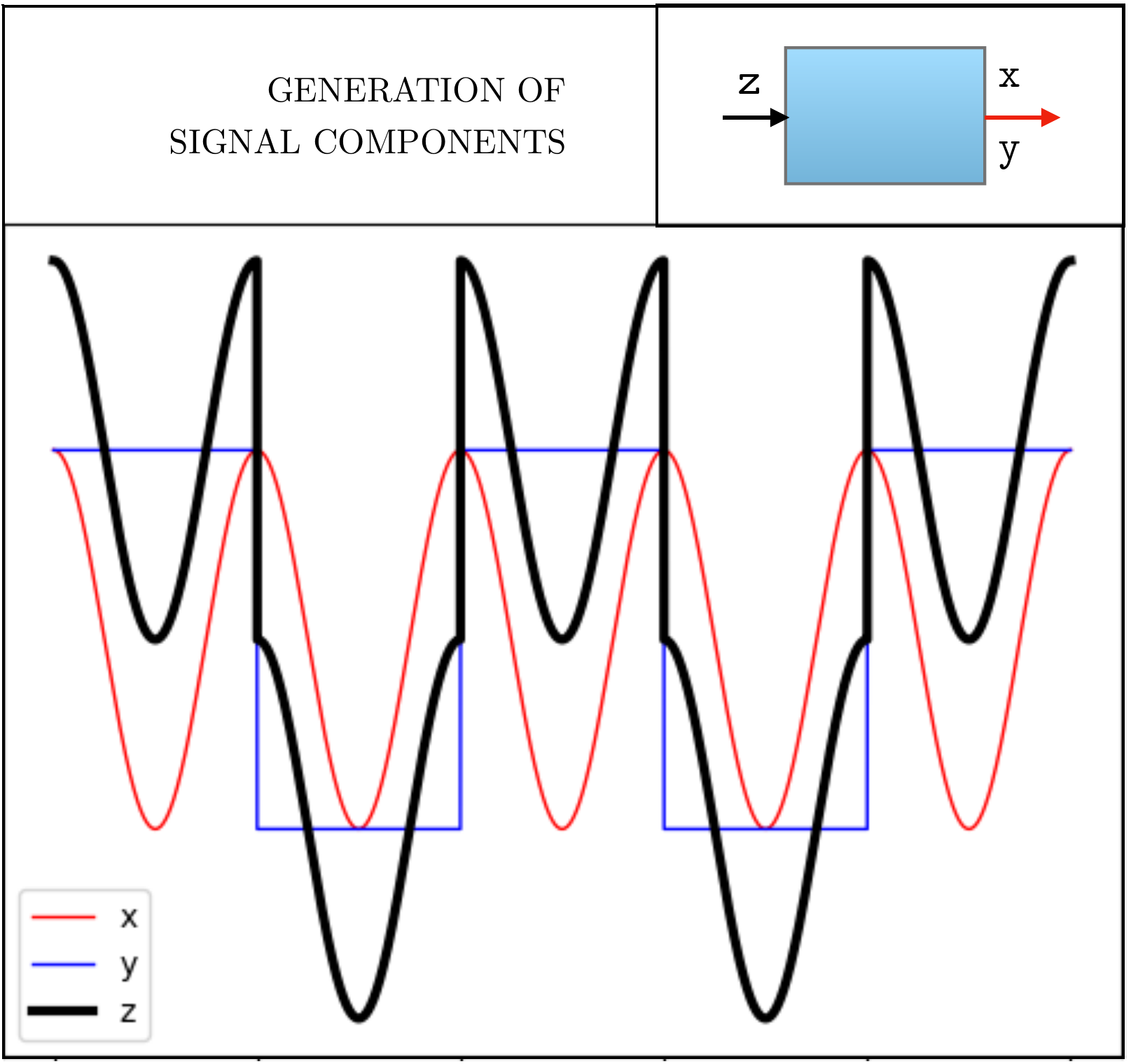}
    %\caption{\small The same generation task as described in the left side figure,
    %withe additional feature of selecting one of the two components of 
    %signal $z$ which comes from the sum of signals $x$ and $z$.
    %}
    \label{fig:figure2}
  \end{minipage}
  \caption{\small \textit{Left side}: Autonomous generation based on the emulation of 
    a previously presented signal (training). In (a) there is a neat separation
    between learning and test, whereas in (b) we consider overlapping of the 
    two phases. The prediction is based on a fixed-length window. The 
    extreme formulation corresponds with a window composed of one sample only.\\
    \textit{Right side}: The same generation task as described in the left side figure,
    with the additional feature of predicting one of the two components
    ${\tt x}$ or ${\tt y}$ of the signal ${\tt z}$ (black signal in the figure).}
\end{figure}

\noindent\textsc{Signal Generation by Target Emulation}\\
Here we propose a classic generation task where the purpose of an agent  
is that of a truly autonomous signal generation. 
More formally, we assume that the agent interacts with the environment by acquiring a
discrete signal $z_{\kappa}, \ \kappa \in {\cal K}_{L}$. The agent is expected to act 
over ${\cal K} = [0,\infty)$, which is partitioned 
into ${\cal K}_{L}$ (learning data) and ${\cal K}_{T}$ (test data).
Hence the agent can only access the signal over ${\cal K}_{L}$ and it 
is expected to generalize over ${\cal K}_{T}$. 
The distinctive feature of the collectionless Machine Learning protocol 
is that there is only an on-line acquisition of $z_{\kappa}$, which formally
means that, given any $k \in {\cal K}$ one can only store 
on-line the segment ${\cal S}_{\kappa}$ that is characterized by the condition
$  
\forall \kappa \in {\cal K}: \ \ |{\cal S}_{\kappa}| = n.
$
Notice that the storing of ${\cal S}_{\kappa}$ takes place uniformly for all indexes
of ${\cal K}$.
Hence, the signal defined by the sequence ${\cal K}$ can be acquired on-line
by a buffer of, at most, dimension $n$, which is a constant chosen in advance. 
Clearly, the notion of Collectionless Machine Learning can be ideally expressed 
in the extreme case in which $n=1$ that is referred to as 
\textit{Extreme Collectionless Signal Generation} (ECSG).
Music generation is an example of this generation process. We assume that the music agent listens 
music over a certain interval and, later on, it generalizes by  playing music itself. ECSG also offers a natural framework for modeling  video generation. 
Notice that the \textit{Extreme Collectionless Signal Generation} imposes 
strong constraints on the learning environment. As yet, to the best of our
knowledge, there is no evidence of intelligent agents based on Machine Learning
that can generalize under this constraint. Interestingly, there is plenty of evidence that 
ECSG related protocols can be found in nature.   \\
~\\
\textsc{Signal Generation of its components}\\
A more advance task is that of generating a signal like in the previous case
under the additional requirement that it is a component
of the acquired input. For example, instead of simply learn to generate the music from a guitar, the agent can be asked to generate music coming either from a piano or form a guitar on the basis of the signal deriving from the composition of the two instruments. The agent can also be asked to replicate the previous task and  play both the guitar and the piano. Clearly, in this case we assume that the input is removed at test time. This also suggests the additional task in which the signal component must be generated at test time even in absence of the input.  The specification of the generation mode can be regarded as a general associative map the purpose of which could also be to grasp a sort of \textit{generation style}. \\
~\\
\textsc{Tasks from Unsupervised Learning}\\
We can easily incorporate the spirit of classic unsupervised learning problem
in the framework of Collectionless AI. For example many tasks in computer vision
can naturally be formulated by simply imposing the satisfaction of the
unsupervised condition at any time without data storing. As an example, 
let us consider the task of optical flow. We can impose the classic 
Horn-Schunk's~\cite{HornAI1981} \textit{brightness invariance condition} at any time and ask the agent to learn the estimation of the corresponding optical flow. {In doing so, we promote solutions that carry out the updating the model predictions over time, instead of  estimating the optical field from scratch given a pair of frames, as usually done offline.}

\section{Conclusions}
This paper proposes a new view of AI which is centered around the idea of Collectionless AI . 
The new learning protocol assumes that machines interact in own environment without permission of neither accessing data collections nor of storing information {to re-create the typical conditions of offline learning. Machines are expected to develop their own memorization skills, by abstracting the information acquired from the sensors, that is processed in an online manner.} We argue that this new framework, might open the doors to a new approach to Machine Learning that can contribute to reconcile symbolic and sub-symbolic AI. Moreover, the emergence of the collectionless philosophy can contribute to better understand intelligence processes in nature as well as to open an alternative technological path which is not centered on the privilege of controlling large data collections.

\section*{Acknowledgments}
\vskip -2mm
We thank {Alessandro Betti, Yoshua Bengio, Giuseppe De Giacomo, Maurizio Lenzerini, Vincenzo Lomonaco, Fr\'ed\'eric Precioso, Paolo Traverso, Luciano Serafini, Bernardo Magnini, Marcello Pelillo, and Fabio Roli} for insightful discussions and, especially, for the criticisms that arose on the proposed
formulation of the collectionless principle.

\end{document}